\documentclass[11pt,a4paper]{article}
\usepackage[hyperref]{acl2020}
\usepackage{times}
\usepackage{latexsym}

\usepackage{microtype}
\usepackage{graphics}
\usepackage{graphicx}
\usepackage{amsmath}
\usepackage{enumitem}
\usepackage{color}
\usepackage{booktabs}
\usepackage{multirow}
\usepackage{makecell}

\aclfinalcopy % Uncomment this line for the final submission
\newcommand{\blue}[1]{\textcolor{blue}{#1}}

\title{Local Knowledge Powered Conversational Agents}

\author{Sashank Santhanam\thanks{Work was done during internship at NVIDIA. } \\
  \hspace{-1.5em}Computer Science Department \\
  \hspace{-1.5em}UNC Charlotte \\
  \hspace{-1.5em}\texttt{ssantha1@uncc.edu} \\
  \And
   Wei Ping  \\
   NVIDIA \\
   \texttt{wping@nvidia.com} \\
  \And
  \hspace{1.2em}Raul Puri \\
  \hspace{1.2em}OpenAI \\
  \hspace{1.2em}\texttt{raulpuric@berkeley.edu} \\
  \AND  
  \hspace{-1.5em}Mohammad Shoeybi \\
  \hspace{-1.5em}NVIDIA \\
  \hspace{-1.5em}\texttt{mshoeybi@nvidia.com}
  \And
  Mostofa Patwary \\
  NVIDIA \\
  \texttt{mpatwary@nvidia.com}
  \And
  \hspace{1.5em}Bryan Catanzaro \\
  \hspace{1.5em}NVIDIA \\
  \hspace{1.5em}\texttt{bcatanzaro@nvidia.com}
  }
\date{}

\begin{document}
\maketitle

\begin{abstract}
State-of-the-art conversational agents have advanced significantly in conjunction with the use of large transformer-based language models. However, even with these advancements, conversational agents still lack the ability to produce responses that are informative and coherent with the local context.
In this work, we propose a dialog framework that incorporates both local knowledge as well as users' past dialogues to generate high quality conversations.
We introduce an approach to build a dataset based on \emph{Reddit} conversations, where outbound URL links are widely available in the conversations and the hyperlinked documents can be naturally included as local external knowledge. 
Using our framework and dataset, we demonstrate that incorporating local knowledge can largely improve \emph{informativeness}, \emph{coherency} and \emph{realisticness} measures using human evaluations.  In particular, our approach consistently outperforms the state-of-the-art conversational model on the \emph{Reddit} dataset across all three measures. 
We also find that scaling the size of our models from 117M to 8.3B parameters yields consistent improvement of validation perplexity as well as human evaluated metrics. 
Our model with 8.3B parameters can generate human-like responses as rated by various human evaluations in a single-turn dialog setting.

\end{abstract}

\section{Introduction}

One of the biggest challenges in conversational AI and dialog systems is building human-like conversational agents that are capable of generating \emph{realistic}, \emph{informative}, and \emph{coherent} responses, so that users find them engaging and enjoy the ongoing conversation. Traditionally, conversational agents are built using RNN-based \textit{seq2seq} models~\cite[e.g.,][]{vinyals2015neural}. However, these models tend to generate vague and generic responses that are less engaging~\cite[e.g.,][]{li-etal-2016-persona}. 
Recent advances in large-scale language models~\cite{radford2019language, shoeybi2019megatron,raffel2019exploring, brown2020language} have pushed the state-of-the-art in Natural Language Generation~(NLG), paving the way to use transformer-based models~\cite{vaswani2017attention} in end-to-end dialog systems. 

% personalized conversational agents
There have been several efforts~\citep{wolf2019transfertransfo, golovanov2019large} to apply the large-scale language models to build more engaging personalized conversational agents on the supervised Persona-Chat dataset~\citep{zhang-etal-2018-personalizing}. 
These models can produce conversations that adhere to the reference profile facts, but are devoid of unique personality and instead exhibit a mean average style~\citep{boyd-etal-2020-large}. 
Most recently, \citet{boyd-etal-2020-large} introduced a dataset based on conversations from \emph{Reddit} comments and built a conversational agent that conditions on a knowledge base of past reference conversations to model the speaker’s persona. However, it only considers past dialogues and did not use any external knowledge to ground the generations.

% external knowledge for conversational agents
Some previous studies \cite{DBLP:conf/iclr/DinanRSFAW19, qin-etal-2019-conversing, shuster2020image} attempted to improve coherence and informativeness of dialogues by incorporating external knowledge bases~(e.g., Wikipedia articles, images) into the conversational agents. However, the dialogues in these datasets are artificially designed and may not reflect the diversity or quantity of real-world conversations. 

\begin{figure*}[t]
  \centering
  \small
    \includegraphics[width=0.90\textwidth]{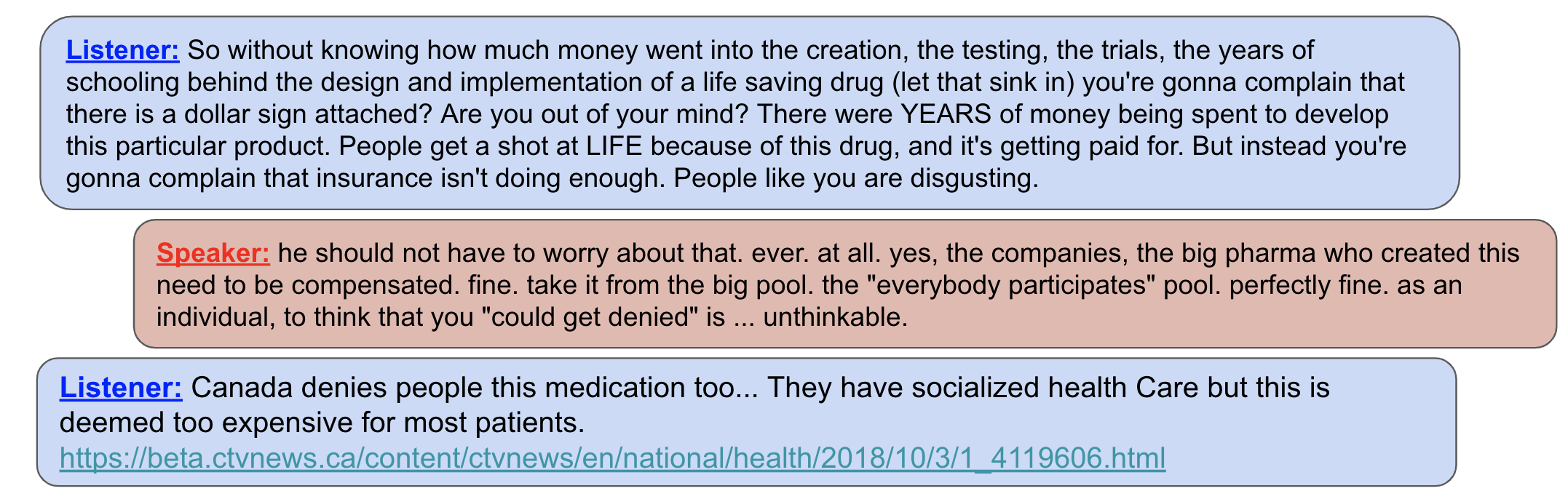}
   \vspace{-.5em}
  \caption{An excerpt from a \emph{Reddit} conversation between a speaker and a listener about a particular topic. As the conversation proceeds, a new piece of evidence is introduced by the listener through an URL.}
  \label{fig:example}
\end{figure*}

In this work, we aim to improve dialogue's coherence and informativeness by incorporating local knowledge in a self-supervised framework for a large, web-scraped persona dataset. We use references to external links in the current dialog as the source for local knowledge. Indeed, local references for external knowledge widely exist in online conversations between humans. For example, we find that during conversations on platforms such as \emph{Reddit}, users often use hyperlinked documents~(e.g., by URLs) as additional pieces of evidence to ground their statements. Consider the example shown in Figure \ref{fig:example}, a small snippet of a conversation between a speaker and a listener, where the listener posts a URL in the last turn. These hyperlinked documents usually contain relevant pieces of information that are closely related to the current conversation.
In spite of that, they were ignored or filtered out by previous work~\citep{zhang2019dialogpt, boyd-etal-2020-large}.

Our primary goal here is to learn a model that is able to generate high quality responses by modeling the past dialogues of the speaker as well as attending to any external document that has been referred to throughout the conversation. To do so, we present a dialog framework that combines the retrieval and generation process together. We build upon \citet{boyd-etal-2020-large} by using \emph{Reddit} comments as our data source, and build an external knowledge base with the user-posted outbound links referenced throughout dialogues. We perform a K-Nearest-Neighbour (KNN) based search to retrieve relevant evidence phrases from the external documents and use them to context prime the model. Recent work by \cite{fan2020augmenting} also incorporates external knowledge into the conversational agents through information retrieval. Unlike their approach that uses Wikipedia, pre-defined images, and dialogue knowledge bases, our work ensures that diverse sources of knowledge are used by performing retrieval from hyperlinked documents introduced in a conversation.
We also find that limiting the search space for KNN to a local knowledge base, rather than a global knowledge base such as Wikipedia, ensures that the most relevant and informative context is retrieved when generating a response. In addition, similar to \citet{boyd-etal-2020-large} we incorporate persona into the responses using user's past dialogues to ensure that the generated response is consistent with the speaker's style of writing and their opinion on certain topics.

In summary, our contributions are as follows: 
\begin{itemize}[itemsep=-1.00pt, topsep=3.5pt, leftmargin=1.0em]
    \item We propose a dialog framework that incorporates both local external knowledge and user's past dialogues to generate high-quality responses.
    \item We present an approach to creating a dataset based on \emph{Reddit} conversations, which uses outbound links in the comments as the external knowledge.
    \item We demonstrate that incorporating the local knowledge consistently improves \emph{informativeness}, \emph{coherency} and \emph{realisticness} measures when compared to ground truth human responses. In addition, our model outperforms the state-of-the-art conversational agent on the \emph{Redddit} dataset~\cite{boyd-etal-2020-large}, as it exploits both external knowledge and user's past dialogues. 
    \item We show that scaling up our model from 117M to 8.3B parameters consistently decreases the validation perplexity from {20.16 to 12.38} based on a vocabulary of 50K BPE subwords~\citep{sennrich2015neural}. In particular, our 8.3B model generates high quality responses on par with human responses in terms of  \emph{informativeness}, \emph{coherency} and \emph{realisticness} evaluations.
\end{itemize}

We organize the rest of the paper as follows. 
We present the framework of our conversational agent in Section~\ref{sec:framework}, and introduce the dataset creation in Section~\ref{sec:dataset}.
We present the experiment and evaluation setup in Section~\ref{sec:experiment}, and report the results in Section~\ref{sec:results}.
We further discuss the related work in Section~\ref{sec:related_work}, and conclude the paper in Section~\ref{sec:conclusion}.

\section{Framework}
\label{sec:framework}
Consider the conversation $\{X_i\}_{i=1}^{n-1}$, where $X_i$ is a turn in the conversation between two or more users. The task is to generate the turn $X_n$ for user $A$~(i.e., the speaker in Figure~\ref{fig:example}) given the current conversation. It is done by using our framework illustrated in Figure~\ref{fig:arch}, which consists of three components:

\begin{figure*}[t!]
  \centering
  \small
  \includegraphics[width=.87\textwidth]{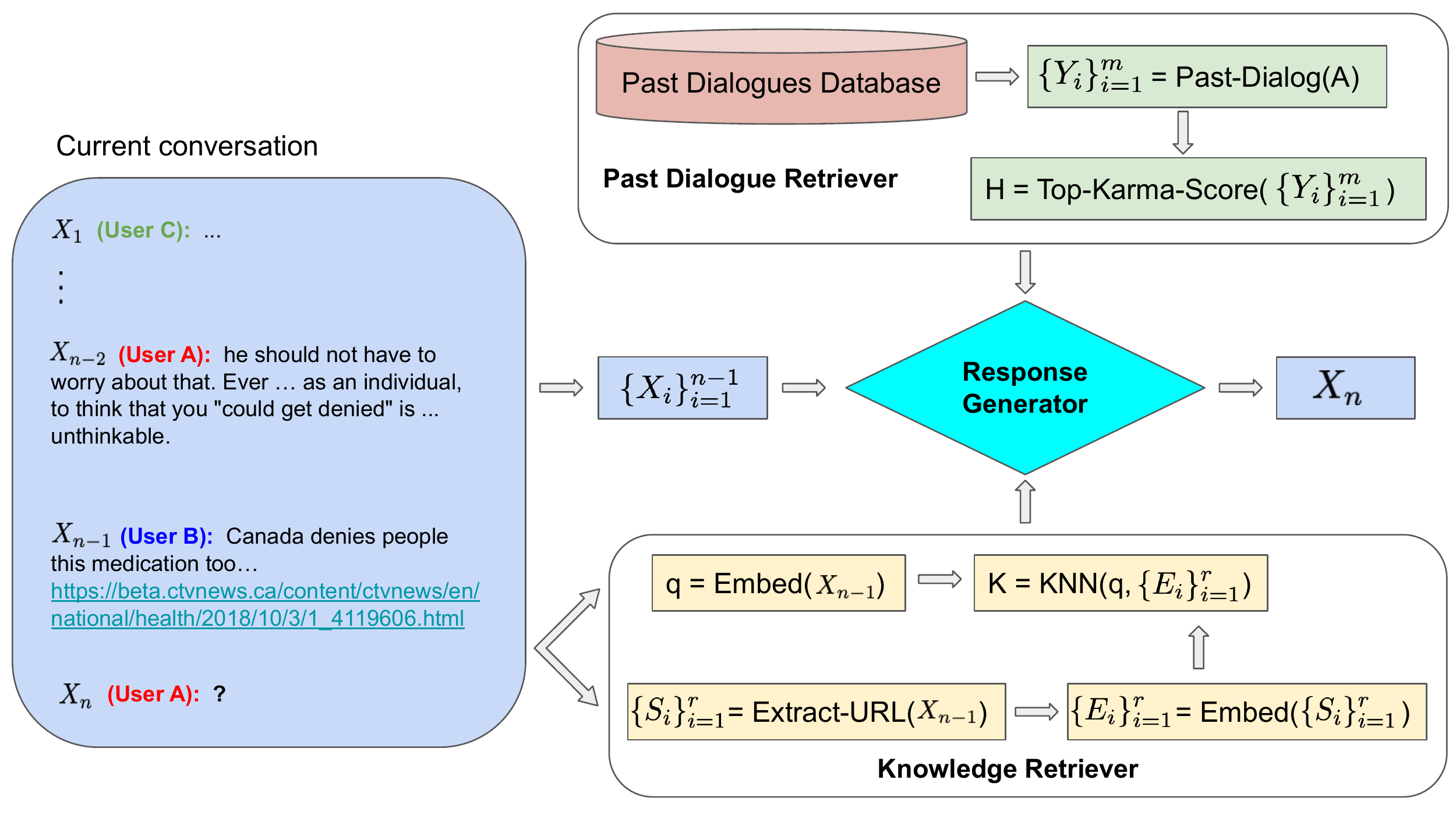}
  \vspace{-.2cm}
  \caption{Architecture diagram of our framework consisting of the following components: (i) Knowledge Retriever: helps retrieve relevant sentences $K$ from the URLs; (ii) Past Dialogue Retriever: retrieves past dialogues $H$ from user $A$ who is generating our response; (iii) Response Generator: a GPT-2 model that is to be finetuned and take the knowledge retrived along with past dialogues and the current conversation as input.}
  \label{fig:arch}
\end{figure*}

\begin{figure}[t]
    \centering
    \hspace{-.5cm}
    \includegraphics[width=8cm, height=3cm]{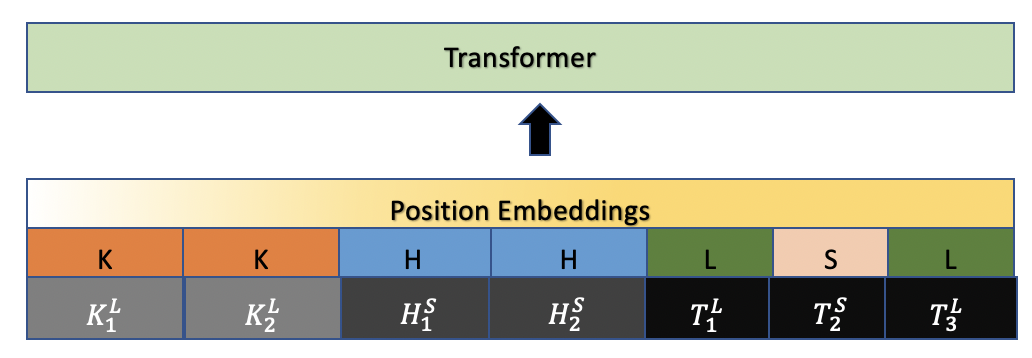}
    \caption{Illustration of input representation to the GPT-2 transformer model for a conversation between a listener ($L$) and speaker ($S$) (to be modeled). Along with the subword embeddings of current conversation  $(T_i)$, the model also receives past dialogues from the speaker's history ($H_i^s$) and most relevant knowledge sentences  ($K_i^L)$ introduced by listener. We also add positional embeddings and token type embeddings K, H, L, S for knowledge, past dialogues, listener, and speaker, respectively. }
    \label{fig:input_representation}
\end{figure}

\begin{itemize}[itemsep=-1.00pt, topsep=3.5pt, leftmargin=1.0em]
    \item \textbf{Knowledge Retriever:} To include external knowledge, we consider $X_{n-1}$, the last turn in the current conversation, and extract the information referenced by the outbound URL links. The extracted knowledge is then divided into sentences $\{S_i\}_{i=1}^{r}$ and each sentence $S_i$ is encoded to a fixed size vector $E_i$ using Universal Sentence Encoder \cite{cer2018universal}, denoted by USE. The knowledge retriever encodes the last turn of the conversation $X_{n-1}$ as query $q$ using the same USE embedding, and performs a cosine similarity search between $\{E_i\}_{i=1}^{r}$ and $q$. Then, it picks $k$ sentences $K = \{S_i\}_{i=1}^{k}$ with the highest similarity scores. We simply set $k=5$ in all experiments.
    In our framework, we pre-compute all the sentence embeddings $\{E_i\}_{i=1}^{r}$ associated with all the external URLs and build a knowledge-base called Ext-Docs which includes all the documents referenced within the dataset.
    
    \item \textbf{Past Dialogue Retriever:} We denote $Y = \{Y_i\}_{j=1}^{m}$ as the past dialogue turns associated with speaker $A$. Note that past dialogues contain all the historical comments made by each user but does not contain any utterances from the current conversation. To retrieve the relevant and high-quality past dialogues emblematic of a user's personality, we follow the heuristic strategy used by \citet{boyd-etal-2020-large}. We sort the past dialogues based on their karma score in \emph{Reddit} (which is the difference between the up-votes and down-votes of a comment) and pick the ones with the highest scores. We denote the retrieved past dialogues as $H$.
    
    \item \textbf{Response Generator:} Traditionally, the goal of the response generation component has been to produce an informative response $X_n$ conditioned on the current conversation turns $\{X_i\}_{i=1}^{n-1}$. However, just incorporating these turns might not provide enough information to produce an informative response \cite{fan2020augmenting, wolf2019transfertransfo}. In order for the response generator to make use of the retrieved knowledge and past dialogues, we concatenate them as part of the conditional input for a left-to-right GPT-2 language model~\cite{radford2019language}, in which the input context size is $1024$ in our experiments. The retrieved knowledge and past dialogue sequences are truncated to a maximum of $250$ tokens each, and the current conversation is allocated a minimum of $524$ tokens in case there is no past dialog for a particular user or outbound URL links in current conversation. 
\end{itemize}

We illustrate the input representation to the GPT-2 response generator in Figure~\ref{fig:input_representation}.
In addition to the positional embedding, we tell the model which sequence of tokens is from the speaker of interest or listeners in the current conversation, which are retrieved external knowledge, and which are speaker's past dialogue. 
This is achieved by adding token type embeddings to the positional encoding and subword embeddings.

\section{Dataset}
\label{sec:dataset}
In this section, we present our dataset creation approach. To create a large-scale dataset for self-supervised learning, we rely on the publicly available archive of \emph{Reddit} comments that has been made available on pushshift.io \footnote{\url{https://files.pushshift.io/reddit/comments/}}. In our work, we use conversations extracted from a subset of months ranging from October 2018 to April 2019. 
We extract conversations as a sequence of turns by traversing through \emph{Reddit}'s comment graph structure. To ensure that the large volume of comments are of high quality, we apply the filtering strategy proposed by  \citet{boyd-etal-2020-large}  and add other conditions to further improve the quality of the conversations. Adding all these filtration rules together, we extract conversations based on the following conditions:
\begin{itemize}[itemsep=-1.00pt, topsep=3.5pt, leftmargin=1.5em]
    \item The conversation has a minimum of 5 turns.
    \item The conversation has a maximum of 15 turns.
    \item At least one turn has minimum karma score of 4 within the conversation.
    \item All turns in the path have at least 3 words.
    \item The conversation shares a maximum of 2 turns with previously extracted paths.
    \item No turns in the path originate from a ``Not Safe For Work'' subreddit.
    \item No user in the conversation is marked as ``Deleted''.
\end{itemize}

We process each month individually in parallel. Once all the conversations were extracted from a specified month, we then extract all the URLs mentioned in each turn of a conversation to create the knowledge base of hyperlinked documents (Ext-Docs knowledge-base). The URLs are filtered out based on an undesirable list of domain names and extensions. We use the two block lists found in the Megatron-LM repository \footnote{\url{https://github.com/NVIDIA/Megatron-LM}}.

Overall, we extracted 48M conversations and found that 10.4\% of the  conversations had used a URL as a piece of evidence in the conversation. To create a more balanced dataset between conversations that use no URLs and conversations that use URLs, we downsample the conversations with no URLs. After downsampling, we ended up with a total of {1,585,875} conversations where {1,232,244} of these conversations had no URLs and {353,631} conversation had used URLs. We further split the filtered dataset with a 80-10-10 ratio to create the training, validation, and test sets.

Additionally, we precomputed all the past dialogues made by users across the time span of our dataset (2018-10 to 2019-04) and stored them. In the final dataset, we had {593,734} unique users and on average each user had around {21.13} historical comments.

\section{Experiments}
\label{sec:experiment}
In this section, we present the experimental setup as well as the automatic evaluation metrics and human evaluation metrics we used in the experiments.

\subsection{Experimental Setup}

We implement our models using the Megatron-LM repository~\cite{shoeybi2019megatron}. 
For the majority of our experiments, we use the pretrained GPT-2 model~\cite{radford2019language} with 345M parameters, 24 layers and 16 attention heads. The input utterances, retrieved knowledge and past dialogues are tokenized using byte-pair encoding~(BPE) to reduce vocabulary size \cite{sennrich2015neural}. The vocabulary size is set to 50,262 with the addition of special tokens for demarcating the beginning and ending of past dialogues (\_\_bpd\_\_, \_\_epd\_\_), the beginning and ending of knowledge (\_\_bk\_\_, \_\_ek\_\_), and speaker and listener segment tokens. We use the Adam optimizer with a cosine learning rate decay warmed up over 1\% of total iterations. Overall the model is fine-tuned for 55,000 iterations with a global batch size of 64. 
At the training phase, the input sequences are concatenated along the length dimension, as is a common practice for transformer inputs~\cite{devlin-etal-2019-bert,wolf2019transfertransfo}.   
For decoding, we use nucleus sampling with $p = 0.9$ to generate responses~\citep{holtzman2019curious}.

\subsection{Models}\label{subsec:model_comparision}
We investigate four different models to demonstrate the benefits of incorporating past dialogues and local knowledge:
\begin{itemize}[itemsep=-1.00pt, topsep=3.5pt, leftmargin=1.0em]
    \item \textbf{Baseline (B):~} The simplest of the four models used in our experiments, which is used to establish a baseline. In this model, only the current conversation, i.e., $\{X_i\}_{i=1}^{n-1}$, is provided as input sequence to the response generator.
    Despite its simplicity, it is a strong baseline for response generation as demonstrated by~\citet{zhang2019dialogpt}.
    
    \item \textbf{Baseline + Past Dialogues (B + H)}: This model is the state-of-the-art response generation approach presented by~\citet{boyd-etal-2020-large}. In this model, a heuristic based approach is used to identify the retrieved past dialogues of a speaker, which is then combined with the current conversation. The retrieved past dialogue (denoted as $H$) is concatenated with the current conversation as the input to the response generator.
    
    \item \textbf{Baseline + Knowledge (B + K):~} This setting measures the importance of adding external knowledge for the response generation process. In this model, we combine retrieved knowledge sentences from the external URLs (denoted by $K$), and concatenate them as additional pieces of evidence to the current conversation.
    
    \item \textbf{Baseline + Knowledge + Past Dialogues (B + K + H)}: This setting measures the importance of incorporating both external knowledge and retrieved past dialogues for the response generation process. In this model, we combine retrieved knowledge sentences $K$ from the external URLs and the retrieved past dialogues $H$ from user that is being modeled. We concatenate them as additional pieces of evidence to the current conversation.
\end{itemize}

\subsection{Automated Metrics}
Automatic evaluation for the quality of generated responses is still an active area of research for open-domain conversation. 
Previous work have used metrics such as BLEU~\cite{papineni2002bleu}, METEOR~\cite{banerjee2005meteor}, ROUGE~\cite{lin2004rouge} from machine translation and text summarization~\cite{dialogue-eval} tasks, although several works have demonstrated that they don't correlate well with human judgments for open-ended tasks such as dialogue~\citep{liu2016not}. 
In this work, we report BLEU score following established reporting practices.
We also report the perplexity (PPL) on the validation set as a measure to compare different models, which was found to correlate with fluency in generations in previous study.

\subsection{Human Evaluation}
Human evaluation is viewed as the most effective way for evaluating the quality of generated text. Traditionally, human evaluation is conducted through the use of Likert scales~\citep{likert1932technique} or continuous scales as the primary experiment design. However, prior research has shown that the usage of Likert scales affects the quality of ratings obtained from the human annotators~\cite{novikova-etal-2018-rankme, santhanam-shaikh-2019-towards}, and the usage of continuous scales such as magnitude estimation is prone to cognitive bias~\cite{SashankCHI}. 
To avoid these issues, we provide pairs of conversations side by side with the last turn generated by either the model or the human and ask the annotator to choose between the two. We also provided a tie option. Overall, we randomly sample 100 conversations from the test set for our evaluations. The annotators are asked to evaluate the quality of the responses according to the following metrics:
\begin{itemize}[itemsep=-1.00pt, topsep=3.5pt, leftmargin=1.0em]
    \item \textbf{Informativeness} measures whether the response from the speaker is informative for listeners (i.e. contains more detailed information). 
    %The human workers were asked to determine which of the last turn from the two conversations is more informative.
    \item \textbf{Coherence} measures whether the response from the speaker matches the topic and discussion from the earlier context of the conversation. %The human workers were asked to determine which of the last turn from the two conversations is more coherent.
    \item \textbf{Realistic} measures whether the response from the speaker looks like a response from a real human instead of a bot. %The human workers were asked to determine which of the last turn from the two conversations is more human.
\end{itemize}
The Mechanical Turk user interface for annotation is provided in Appendix~\ref{sec:mturk_user_interface}.
We utilize 5 unique workers per example in our evaluations. To obtain high quality human labels from native English speakers, the workers are required to reside in the United States and have a Human Intelligence Task~(HIT) approval rate greater than or equal to $95\%$.
We explicitly state in the instructions that payment is contingent on raters spending at least 25 seconds per assignment. 
We tried to filter the inexperienced raters based on their past \emph{Reddit} use as in previous study~\cite{boyd-etal-2020-large}, but we found this is less effective as the raters tend to select the maximum hours we provided in our survey.

\section{Results}
\label{sec:results}
In this section, we report the results of automatic and human evaluations detailed in previous section.
\subsection{Automated Metrics}
Table \ref{automated_metrics} provides a comparison of the different models used in our experiments. We find that compared to the baseline model (B), the addition of knowledge or past dialogue reduces the validation perplexity. In particular, adding past dialogue information can improve the perplexity significantly. 
We also notice that the best perplexity is achieved by adding both retrieved knowledge and past dialogues as additional pieces of evidence.  
The BLEU score degrades when we add knowledge and past dialogue separately, but slightly improves as we incorporate them together. As we will demonstrate in human evaluation results, these BLEU scores don't correlate well with human judgements. We don't report BLUE score further.

We also performed ablation studies on the best performing model~(B + K + H) for various model sizes. Table~\ref{scaledup} gives the different configurations of models that were trained. We find that validation perplexity drops significantly as we increase the size of the models. 
These results are consistent with prior studies~\citep[e.g.,][]{shoeybi2019megatron}.

\begin{table}[t!]
\centering
\begin{tabular}{l|cc}
\Xhline{1pt}
Models  & Val PPL($\downarrow$)  & BLEU($\uparrow$)   \\  \hline
B & 18.12    & 15.3                       \\ 
B + K & 18.10  & 14.1                           \\ 
B + H & 16.84   & 14.0                          \\ 
B + K + H     & \textbf{16.83}       & \textbf{15.4}           \\  \Xhline{1pt} 
\end{tabular}
\caption{Automated metrics results (Val PPL and BLEU) on the test set obtained by fine-tuning the 345M model with different experimental settings. {\bf B}: stands for baseline model that only exploits current dialog context. {\bf H}: stands for the heuristic approach for retrieving past-dialogues. {\bf K}: stands for retrieval of knowledge. $\uparrow$ means the number is the higher the better, and $\downarrow$ means the number is the lower the better.}
\label{automated_metrics}
\end{table}
\begin{table*}[t!]
\centering
\begin{tabular}{l|cccc|c}
\Xhline{1pt}
Model & Hidden size & Layers  & Attention heads & \#Parameters & Val PPL($\downarrow$) 
\\ \hline
B + K + H   & 768      & 12    &   12      & 117M   & 20.16 
\\ 
B + K + H    & 1024    & 24    &       16   & 345M      & 16.83 
\\ 
B + K + H & 1536    & 40      &          16   & 1.2B   & 14.57 
\\ 
B + K + H & 3072    & 72      &          24   & 8.3B   & {\bf12.38}  
\\ \Xhline{1pt}
\end{tabular}
\caption{Scaled up results for our best performing model~(B + K + H). $\downarrow$ means the lower value is the better.}
\label{scaledup}
\end{table*}

\subsection{Human Evaluation}
\begin{table*}[t!]
\centering
%\vspace{0.2em}
\begin{tabular}{l !{\vline width 1pt} c|c|c !{\vline width 1pt} l}
\Xhline{1pt}
\textbf{Source X}  & \textbf{Informativeness} & \textbf{Coherency} & \textbf{Realisticness} & \textbf{Source Y} \\ \hline
B~(345M) & 28\% - 20\% - \blue{52\%} & 29\% - 22\% - \blue{49\%} & 31\% - 33\% - \blue{36\%}  & Human 
\\
B + K~(345M) & 31\% - 26\% - \blue{43\%} & 30\% - 27\% - \blue{43\%} & 26\% - 36\% - \blue{38\%} & Human \\

B + H~(345M)  & 29\% - 31\% - \blue{40\%} & 26\% - 33\% - \blue{41\%} & 29\% - 21\% - \blue{50\%} & Human 
\\ 

{\bf B + K + H}~(345M)  & 34\% - 29\% - \blue{37\%} & 29\% - 33\% - \blue{38\%} & 26\% - 39\% - \blue{35\%} & Human \\ \hline

\Xhline{1pt}
\end{tabular}
\caption{Pairwise comparison results~(X wins - Ties - \textcolor{blue}{Y wins}) between 345M models and human-generated text using Mechanical Turk. B: stands for baseline model that only exploits current dialog context. R: stands for retrieval for past-dialogues.  H: stands for the heuristic approach for past-dialogues. K: stands for retrieval for knowledge. 
To make relative comparison between models, we highlight the last columns of the results~(best viewed in color).
They indicate the percentages of cases that the models are outperformed by human, which are the lower the better.}
%\vspace{-.1cm}
\label{tab:human_eval_345M}
\end{table*}
\begin{table*}[t!]
\centering
%\vspace{-0.2cm}
%\vspace{0.2em}
\begin{tabular}{l !{\vline width 1pt} c|c|c !{\vline width 1pt} l}
\Xhline{1pt}
\textbf{Source X}  & \textbf{Informativeness} & \textbf{Coherency} & \textbf{Realisticness} & \textbf{Source Y}
\\ \hline
{B + K + H}~(345M)~\hspace{-.6em}  & 41\% - 33\% - 26\% &  29\% - 44\% - 27\%  & 40\% - 24\% - 36\% & B + H~(345M)~\hspace{-.6em} \\ 
{B + K + H}~(1.2B)~\hspace{-.6em}  & 37\% - 36\% - 27\% &  34\% - 36\% - 30\%  & 37\% - 40\% - 23\%  & B + K + H~(345M)~\hspace{-.6em} \\ 
{B + K + H}~(8.3B)~\hspace{-.6em}  & 38\% - 31\% - 31\% &  35\% - 35\% - 30\%  & 33\% - 39\% - 28\%  & B + K + H~(1.2B)~\hspace{-.6em} \\ 
{\bf B + K + H}~(8.3B)~\hspace{-.6em}  & 38\% - 22\% - 40\% &  37\% - 26\% - 37\%  & 41\% - 19\% - 40\%  & Human \\
\Xhline{1pt}
\end{tabular}
\caption{Pairwise comparison results~(X wins - Ties - Y wins) between our best performing model~(B + K + H) and a state-of-the-art model on \emph{Reddit} data~(B + H)~\citep{boyd-etal-2020-large}. 
We also include pairwise comparisons with different model sizes.
We find the larger model always outperforms smaller one across all three metrics.
In particular, Our model with 8.3B parameters can generate high quality responses on par with human responses.}
%\vspace{-.1cm}
\label{tab:human_eval_scale-up}
\end{table*}

We report the human evaluation results for different models in Table~\ref{tab:human_eval_345M}. 
Specially, we compare the generated responses from these models to human responses. 
To make relative comparisons between models, we highlight the last column of the results~(X wins - Ties - \blue{Y wins}) ; they indicate the percentages of cases where the models were outperformed by humans, thus the lower the better.
We draw the following observations:
\begin{itemize}[itemsep=-1.00pt, topsep=4.5pt, leftmargin=1.0em]
    \vspace{-0.4em}
    \item Adding external knowledge significantly improves the informativeness and coherency metrics for both the baseline model~(B vs. B + K), and previous state-of-the-art model~(B + H vs. B + K + H).
    \vspace{-0.4em}
    \item Incorporating past dialogues also improves both the informativeness and coherency measures for baseline models~(B) and (B + K).
    \vspace{-0.4em}
    \item Our model~(B + K + H) outperforms others, including the state-of-the-art model~(B + H) on the \emph{Reddit} dataset~\citep{boyd-etal-2020-large},  in terms of informativeness, coherency and realistic measures.
\end{itemize}

In Table~\ref{tab:human_eval_scale-up}, we perform comparison between models, including pairwise comparison between our method~(B + K + H) and previous state-of-the-art model~(B + H) for this task. We also scale our model up to 8.3 billion parameters and report the human evaluations results in Table~\ref{tab:human_eval_scale-up}. 
We find the consistent improvements of all evaluated metrics when we increase the size of the model. Noticeably, our 8.3 billion model can generate responses with quality comparable to humans in terms of informativeness, coherency and realistic metrics.

\subsection{Case Study}
Table ~\ref{tab:generated_example} displays a conversation between a speaker and listener where the last turn of the conversation is generated by the model. We also show the top two retrieved sentences from the external URL that is used to generate the response. From the generated response, our model is able to make use of the relevant spans of knowledge such as ``The men training for less than 3 months, on average, squatted 102kg (225lbs)'' and ``The men training for less than 3 months, on average, benched 85kg (185-190lbs)''. 
More samples from our model are provided in Appendix~\ref{sec:generated_samples}.

\begin{table*}[h]
\small
\centering
\begin{tabular}{ll}
\toprule
& Conversation 1     
\\ 
\toprule
Listener            & \begin{tabular}[c]{@{}l@{}}I literally progressed linearly (with SL 5x5) to a 2 plate bench and 3 plate squat in roughly 3 months while \\ deployed to Iraq and  eating roughly at caloric maintenance (not by own volition.)Went from being unable \\ to bench 185.\end{tabular}           
\\ \midrule
Speaker                                                   & \begin{tabular}[c]{@{}l@{}}You better get into professional sports then because apparently you're in the top 0.001\% of genetics. Not sure \\how you expect anyone to believe that not only did your lifts increase in probably the 99th percentile in terms \\ of progress, but you also did it without eating enough food to build muscle. \\
LMAO look around you goober. People constantly post their progress here and it's nowhere near what you're \\ describing. Here's a post from a day ago where it took someone 7 months to go from  a 65 pound bench to 135. \\ Go look at the progress posts in /r/fitness. If anyones misinformed on what the average new lifter can expect \\ in  terms of progress, it's you. 
You suggesting that is normal to  add 5 pounds to the bar every session' to hit \\ 2/3/4 plate lifts is laughable. The fact you said you did it eating at maintenance is even more hilarious.  \\ 
If you posted in /r/fitness what you just posted here, you'd get laughed out of that sub-reddit. \\ It's not even close to realistic to expect someone to LP to a 315 pound squat. Like not even close. \\
You also completely ignored the fact you contradicted yourself by calling me a beginner yet in the same breath \\you say that once you can't LP anymore, you're an intermediate.
Here's an article on average weight lifting \\ progression. Not only are the numbers you just described far above what the average person can expect \\ in a year of training, you supposedly did it in 3 months while eating in caloric maintenance. \end{tabular} 
\\ \midrule
Listener                                                    & \begin{tabular}[c]{@{}l@{}}\url{https://www.strongerbyscience.com/realistic-training-goals/} \\

16\% of lifters training for 3-6 months reached a 264lb bench and 373lb squat. Cheers!\end{tabular}                                                                                                                                                           \\ \midrule

\begin{tabular}[c]{@{}l@{}}Speaker \\ (Model)\end{tabular} 
& 
\begin{tabular}[c]{@{}l@{}}If you read your own article, you'd notice the vast majority of people hit a set of 1-2 plates in 3-6 months. \\ The average raw beginner could easily hit a {\bf$\sim$200 pound squat} and a {\bf 190 pound bench within 3 months} of \\ starting lifting, and that's assuming they're eating the bare minimum to build muscle.\\ \\ \textcolor{blue}{Retrieved Knowledge 1:} The men training for {\bf less than 3 months}, on average, {\bf squatted} 102kg ({\bf 225lbs}),\\ and the men training for between 3 and 6 months squatted 132kg (290lbs) on average.\\ \textcolor{blue}{Retrieved Knowledge 2:} The men training for {\bf less than 3 months}, on average, {\bf benched} 85kg (185-{\bf 190lbs}), \\ and the men training for between 3 and 6 months benched 96kg (210lbs) on average, for a difference of \\ about 3.4kg (7-8lbs) per month. \end{tabular}                              \\ \toprule                                                                                                                              
\end{tabular}
\caption{Conversation between a Speaker and a Listener where the last turn of the conversation is the response generated by our model which makes use of knowledge introduced by the listener through the URL. The top retrieved sentences from the document are also shown with the generated response. The generated response is coherent with the conversation context and consistent with the retrieved knowledge.}
\vspace{-.1cm}
\label{tab:generated_example}
\end{table*}

\section{Related Work}
\label{sec:related_work}
{\bf Transformer Language Models~}
Large-scale transformer-based language models such as GPT \citep{radford2018improving}, GPT-2 \citep{radford2019language}, XLNet \citep{yang2019xlnet}, BERT ~\cite{devlin-etal-2019-bert} have achieved state-of-the-art performance across several downstream NLP tasks. Further research has shown that increasing the sizes of the models~\cite{shoeybi2019megatron, raffel2019exploring,brown2020language} largely improves the quality of generated text and the performance of downstream tasks.
In this work, we also demonstrate that scaling up model size largely improves performance in retrieval augmented dialog modeling. \\

\vspace{-.6em}
\hspace{-1em}{\bf Dialog Modeling~}
Conversational agents are of great importance for a large variety of applications and can be broadly grouped into two categories, namely, (1) \emph{closed-domain} goal-oriented systems that help users with a particular task at hand, (2)~\emph{open-domain} conversational agents that engage in a free form conversation with a human. The latter is also referred to as chit-chat models or chatbots.
In this work, we focus on the open-domain chatbot, which has recently benefited from using the large \emph{seq2sq} transformer architecture~\citep{adiwardana2020towards, roller2020recipes}.

An analysis of the recent progress on open domain conversational agents has shown these agents to be incapable of holding engaging and consistent conversation \cite{roller2020open, huang2020challenges}. Consistency in conversational agents is crucial to gain confidence and trust \cite{huang2020challenges}. One form of consistency is to exhibit a consistent personality that helps have more human-like conversation \cite{zhang-etal-2018-personalizing, li-etal-2016-persona}. Recent work using large pre-trained transformers have shown promising results for personalized dialogue systems on the PERSONA-CHAT dataset \cite{wolf2019transfertransfo, golovanov2019large}. However, this dataset is synthetic. To overcome the drawbacks of this dataset, researchers have used the \emph{Reddit} dataset as a more natural way of building personalized dialogue systems \cite{mazare-etal-2018-training, zhang2019dialogpt, boyd-etal-2020-large}. 

\vspace{-.3em}
Another important trait of conversational agents is their ability to produce informative responses based on external knowledge \cite{roller2020open, DBLP:conf/iclr/DinanRSFAW19}. One way in which external knowledge has been incorporated is by conditioning the model on a set of facts provided in the dataset \cite{ghazvininejad2018a, qin-etal-2019-conversing}. Other approaches for incorporating knowledge is through the use of a retriever based on TF-IDF \cite{gopalakrishnan2019topical, DBLP:conf/iclr/DinanRSFAW19} or KNN \cite{fan2020augmenting}. Our approach also uses a KNN search on external knowledge, but one key difference to~\citet{fan2020augmenting} is that we only do retrieval of relevant  information from  the local hyperlinked document, which ensures the most relevant and informative context is retrieved.

\section{Conclusion}
\label{sec:conclusion}
In this work, we tackle the task of generating informative responses that are coherent with local context for end-to-end dialogue systems through a combination of retrieval and generation process. 
We provide a data collection pipeline from \emph{Reddit} platforms that could facilitate future research. 
Traditionally, knowledge bases such as Wikipedia articles were predominantly used as the source of information to condition the conversational agents.
In contrast, our work exploits the hyperlinked documents introduced during conversations to ground the generated responses. 
We demonstrate that our approach of using retrieved sentences from the external documents and combining that with the past dialogues of the speaker can generate more informative, coherent, and realistic responses in terms of human evaluations.
In the future, we intend to leverage the learnable retrieval approaches such as REALM~\cite{DBLP:journals/corr/abs-2002-08909}, or RAG~\cite{lewis2020retrieval} to improve the retrieval process on the both the knowledge and past dialogues.

\newpage
\bibliography{acl2020}
\bibliographystyle{acl_natbib}

\newpage

\onecolumn
\appendix

\section{Mechanical Turk Setup}
\label{sec:mturk_user_interface}
\begin{figure}[!h]
  \centering
  \small
    \includegraphics[width=1.0\textwidth]{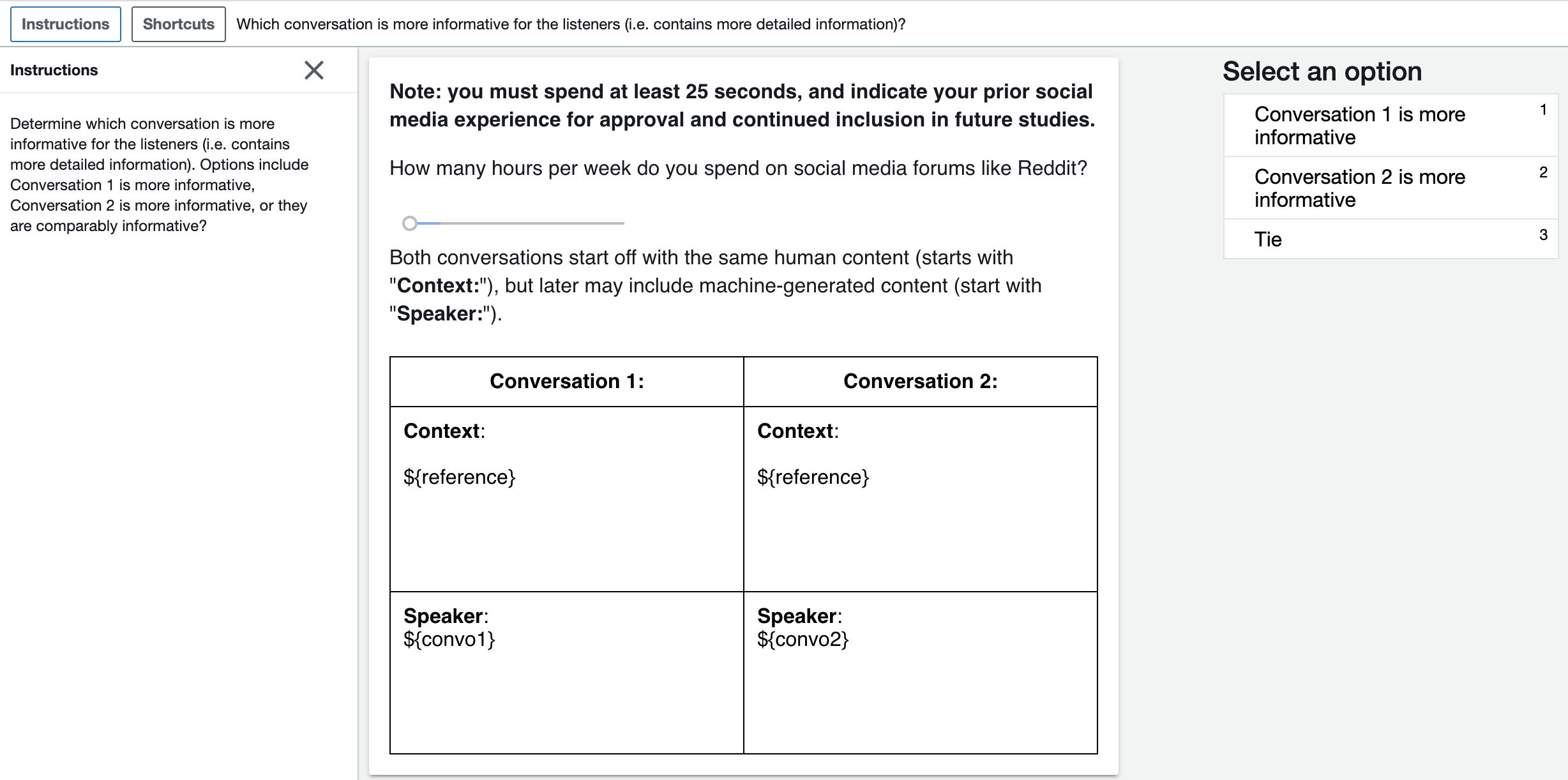}
  \caption{Mechanical Turk interface for informativeness evaluation.}
  \label{fig:mturk_interface_informative}
\end{figure}

\begin{figure}[!h]
  \centering
  \small
    \includegraphics[width=1.0\textwidth]{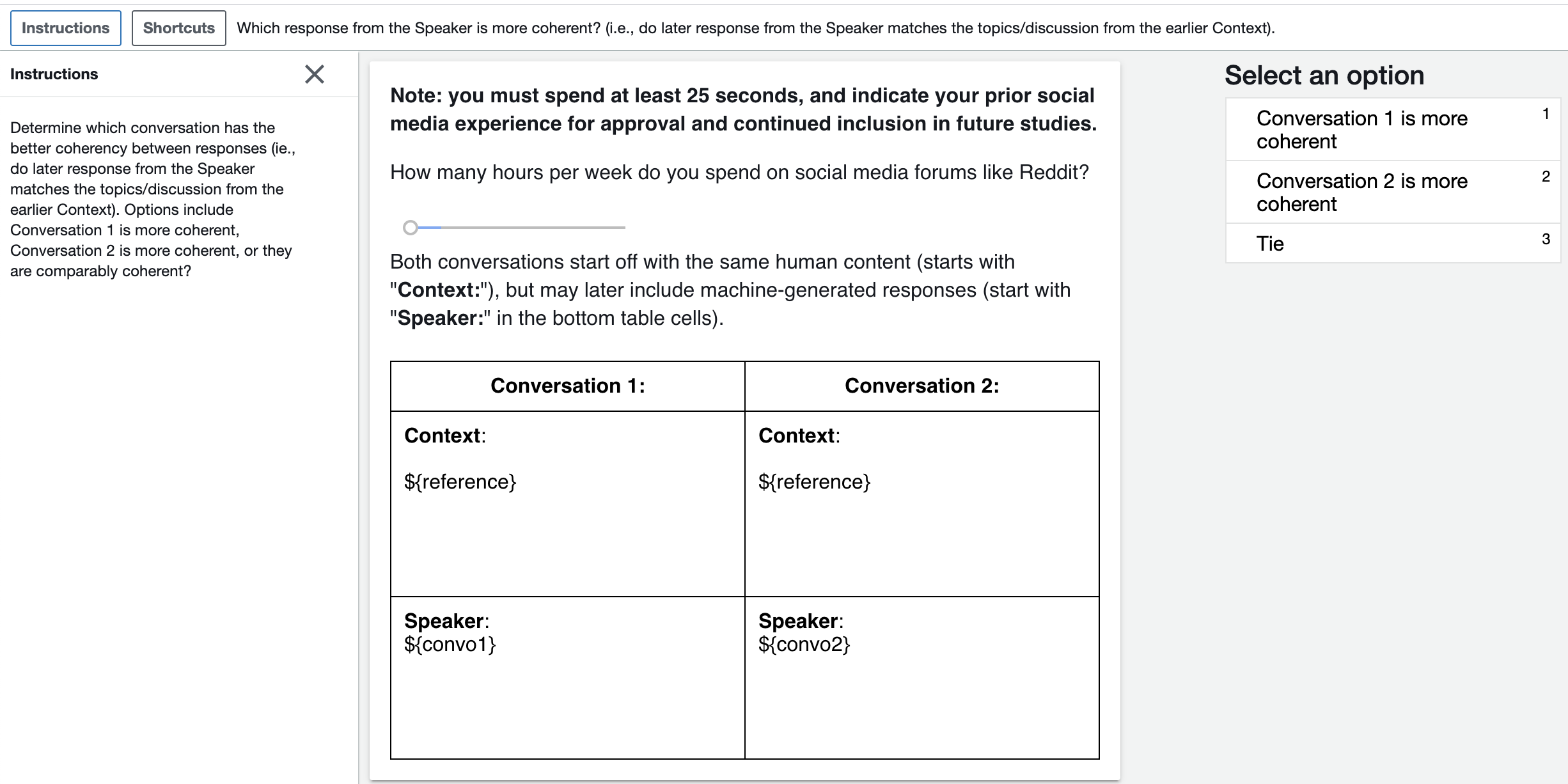}
  \caption{Mechanical Turk interface for coherency evaluation.}
  \label{fig:mturk_interface_coherent}
\end{figure}

\begin{figure*}[!h]
  \centering
  \small
    \includegraphics[width=1.0\textwidth]{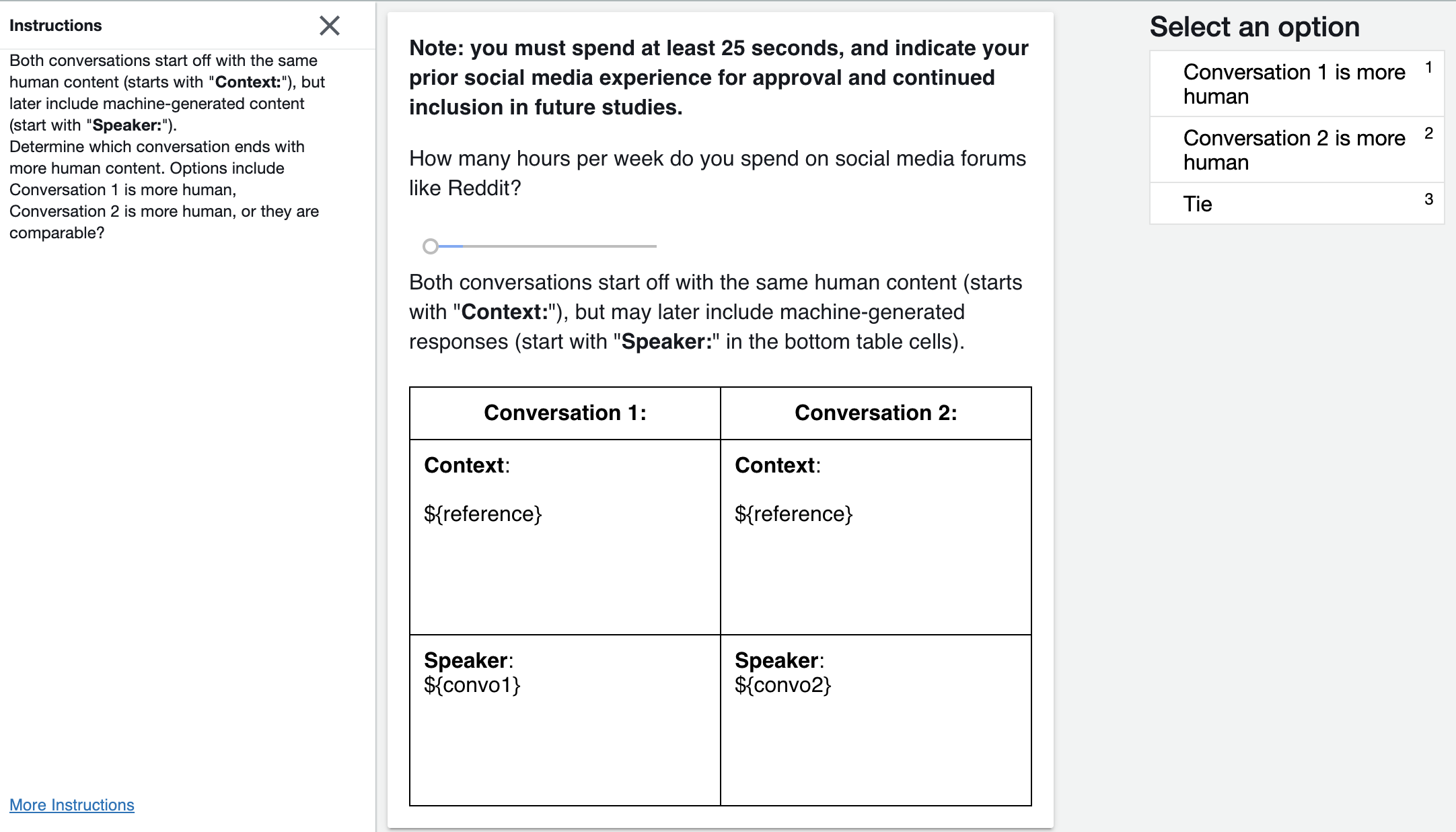}
  \caption{Mechanical Turk interface for realisticness evaluation}
  \label{fig:mturk_interface_realistic}
\end{figure*}

To conduct pairwise comparison,  both conversation 1 and 2 start with the same context from real human conversation.
Then, the last turn from the speaker may include model-generated responses.
One can find the Mechanical Turk interfaces and the detailed instructions for \emph{informativeness} evaluation in Figure~\ref{fig:mturk_interface_informative}, \emph{coherency} evaluation in Figure~\ref{fig:mturk_interface_coherent}, and \emph{realisticness} evaluation in Figure~\ref{fig:mturk_interface_realistic}.

\newpage
\newpage

\vspace{4cm}
\section{Generated Samples from Model}
Table \ref{tab:generated_example1} to ~\ref{tab:generated_example4} are different samples that have been generated by our model. In each of these examples, we show the capability of our model to produce informative responses based on the content that is retrieved by the knowledge retriever.
\label{sec:generated_samples}
\begin{table*}[h]
\small
\centering
\begin{tabular}{ll}
\toprule
& Conversation 2                                                                                                                                                                                                                                                         \\ \toprule
Listener                                                   & \begin{tabular}[c]{@{}l@{}}"Thank god they don't have a choice" hmm ... said no one ever. If you would have been a tad \\ healthier,  you'd have been bankrupt. You're just sick enough. God bless America.\end{tabular}                                                                                                                                                                                                                                                                                                                                                                    \\ \midrule
Listener                                                   & \begin{tabular}[c]{@{}l@{}}So without knowing how much money went into the creation, the testing, the trials, the years\\ of schooling  behind the design and implementation of a life saving drug (let that sink in) you're \\ gonna complain that there  is a dollar sign attached? Are you out of your mind? There were \\ YEARS of money being spent to develop this  particular product. People get a shot at \\ LIFE because of this drug, and it's getting paid for. But instead you're  gonna complain\\ that insurance isnt doing enough. People like you are disgusting.\end{tabular} \\ \midrule
Speaker                                                    & \begin{tabular}[c]{@{}l@{}}he should not have to worry about that. ever. at all. yes, the companies, the big pharma \\ who created this need to be compensated. fine. take it from the big pool. the "everybody \\ participates" pool. perfectly fine. as an individual,  to think that you "could get denied" is \\ ... unthinkable.\end{tabular}                                                                                                                                                                                                                                             \\ \midrule
Listener                                                   & \begin{tabular}[c]{@{}l@{}}Canada denies people this medication too... They have socialized health Care but this is \\ deemed too expensive for most patients.\\  \url{https://tinyurl.com/y5hepduw}\end{tabular}                                                                                                                                                                                                                                                                                                                          \\ \midrule
\begin{tabular}[c]{@{}l@{}}Speaker \\ (Model)\end{tabular} & \begin{tabular}[c]{@{}l@{}}That article is also dated. what are you on about? The drug itself was deemed too \\ expensive for most, and the government mandated that certain patients receive that \\treatment.\\ \\ \textcolor{blue}{Retrieved Knowledge 1:} ``a medicine considered among the most  expensive in the world.”\\ \textcolor{blue}{Retrieved Knowledge 2:} ``provide the drug to the most urgently affected patients  in other groups"\end{tabular}         \\ \bottomrule                                                                                                                                                        
\end{tabular}
\caption{Additional generated sample from our model as in Table~\ref{tab:generated_example}.}
\vspace{-.1cm}
\label{tab:generated_example1}
\end{table*}
%
%
%

%
%
%
%Example 3
\begin{table*}[h]
\small
\centering
\begin{tabular}{ll}
\toprule
& Conversation 3                                                                                      \\ \toprule
Listener                                                   & \begin{tabular}[c]{@{}l@{}}I think you've actually forgotten the initial point you were trying to make. Go back and read the first comment \\ you replied to. It said Neither of pulisic and kovacic were wanted by Sarri 

Then you contradicted that \\ with “Yes, Kovacic is only on loan, of course. I'd like very much that Kovacic will stay with us.” 

A comment \\ made by Sarri 8 months later during which time we lost Cesc, failed to secure his replacement which consequently \\ left Sarri with only Kovacic as the only suitable option.

But we already had 6 midfielders in August, 7 if you count \\ Bakayoko. Why would the board add another one if Sarri didn't want him? 

How is this relevant? \\Baka was always going to go out on loan, Ampadu is a kid, and Drinkwater was never in Sarri's plan to \\ begin with. None of them would've worked as Jorginho's backup anyway.

Sarri might've been okay \\ with Kovacic's transfer back in August but that doesn't mean he wanted Kovacic specifically \\ AT THE TIME, same with Pulisic. It's pretty much agreed upon by the 2 fanbases that Kovacic \\was used to smoothen the deal. \\

\url{https://tinyurl.com/ya2l2323} \\

\url{https://tinyurl.com/y256kld5} \\ \end{tabular}                                                                                                                                                                                                                                                                                                                                                                    \\ \midrule
                                                                           
\begin{tabular}[c]{@{}l@{}}Speaker \\ (Model)\end{tabular} & \begin{tabular}[c]{@{}l@{}} Real Madrid wanted to “sweeten” the Thibaut Courtois-to-Santiago Bernabéu deal with Kovacic, and it looks \\ like Chelsea took the bait. So you really are so in love with Kovacic then huh? 

Baka was always going \\ to go out on loan, Ampadu is a kid, and Drinkwater was never in Sarri's plan to begin with. \\ None of them would've worked as Jorginho's backup anyway. There you go, proof that Sarri \\mdoesn't  want to keep the player for any reason. Interesting. It's great to be able to tell when he's trying \\ to build, whether it's with Kovacic or Pulisic too. \\ \\

\textcolor{blue}{Retrieved Knowledge1:} Madrid wanted to “sweeten” the Thibaut Courtois-to-Santiago Bernabéu deal with \\ Kovacic, and it looks like Chelsea took the bait. \\
\textcolor{blue}{Retrieved Knowledge2:} Real are prepared to allow wantaway Kovacic, who himself skipped a training session \\ with the Spanish club, to join Chelsea on a season-long loan on the proviso they get Courtois. \end{tabular}         \\ \bottomrule                                                                                                                                                        
\end{tabular}
\caption{Additional generated sample from our model as in Table~\ref{tab:generated_example}.}
\vspace{-.1cm}
\label{tab:generated_example2}
\end{table*}

%
%
%
%Example 4
\begin{table*}[h]
\small
\centering
\begin{tabular}{ll}
\toprule
& Conversation 4                                                                                                                                                                                                                                                                                                                                                                                                                                                                                                                                                                              \\ \toprule
Listener                                                   & \begin{tabular}[c]{@{}l@{}}I see where your going and to be fair we don't know until the Olympics get closer, I can see the opposition \\because the money can go to other facets of the city budget but does that have the revenue generator of \\the Olympics, based on opportunity\end{tabular}                                                                                                      \\ \midrule
Speaker                                                   & \begin{tabular}[c]{@{}l@{}}To shrug our shoulders and say "who knows how this will play out" when there are case studies \\ of many previous Olympic Games to learn from? C'mon dude. This isn't just speculation, it's literally every\\  Olympics, this is what happens.\end{tabular} \\ \midrule
Listener                                                    & \begin{tabular}[c]{@{}l@{}}Really you think so ... think about the last few Summer Olympic cities since 1984 when it was LA, \\ Barcelona, Seoul, Sydney, Athens, Beijing, London especially Rio and even Atlanta did not have the \\ same infrastructure that Los Angeles has, to say previous Olympic games, look at the cities.  The venues like the \\ Coliseum in 1984 can still hold the opening ceremonies you don't have to build another like other cities.  \\ The one interesting city in Tokyo will be the best example ... but that's in 2020, so sorry to burst your bubble\\ but it still is speculation and every Olympic games are different based on the community/city ... you know it\end{tabular}                                                                                                                                                                                                                                             \\ \midrule
Speaker                                                   & \begin{tabular}[c]{@{}l@{}}Maybe read up more on this before trying to debate it online, the venue for \\ the opening ceremonies will be the new LA Stadium in Inglewood, which is being built with private \\ NFL money but looks for tax breaks from the city of Inglewood. They will be charging the LA Olympics \\for use of the facility.\end{tabular}                        \\ \midrule

Listener                                                   & \begin{tabular}[c]{@{}l@{}}You're correct but STILL again an already built stadium and hate to break it to you, \\was going to be built regardless of the Olympics.  Los Angeles city wise has a better infrastructure \\than all previous summer Olympic venues since 1984, and I can say is a better city PERIOD so saying \\ it's "every Olympics" is factually wrong\end{tabular}                        \\ \midrule
Speaker                                                   & \begin{tabular}[c]{@{}l@{}}You're focusing too much on whether the stadium was built for the olympics or not. The questions are: \\
who owns the stadium and how much will be charged to the LA Olympics for its use?\\
Will its use also require the use of its onsite vendors, and if so: at what cost? What tax revenue will it generate? \\\end{tabular}                        \\ \midrule

Listener                                                   & \begin{tabular}[c]{@{}l@{}}well I thought you already knew those number since you have historical data literally every Olympics in \\ every city, ready to admit LA is different yet? lol\end{tabular}                        \\ \midrule
Speaker                                                   & \begin{tabular}[c]{@{}l@{}}Do you think other Olympics have never used existing venues before?\end{tabular}                        \\ \midrule

Listener                                                   & \begin{tabular}[c]{@{}l@{}}This isn't just speculation, it's literally every Olympics, this is what happens \& this is what you wrote, \\ Beijing had to build a stadium and there are stories about abandon buildings in Greece \\ \url{https://tinyurl.com/jeuenvx} 
SO how's that EVERY Olympic statement holding up now\end{tabular}                        \\ \midrule

\begin{tabular}[c]{@{}l@{}}Speaker \\ (Model)\end{tabular} & \begin{tabular}[c]{@{}l@{}}From that article you posted:
"For many Greeks who swelled with pride at the time, the Games are now a anger \\ as the country struggles through a six-year depression, record unemployment, homelessness and poverty, with \\ many questioning how the nation has benefited from the multi-billion-dollar event"\\ \\ \textcolor{blue}{Retrieved Knowledge 1:} A decade after the sporting extravaganza, many of its once-gleaming Olympic venues \\ now lie abandoned.\\ \textcolor{blue}{Retrieved Knowledge 2:} For many Greeks who swelled with pride at the time, the Games are now a source of anger \\ as the country struggles through a six-year depression, record unemployment, homelessness and poverty, with many \\ questioning how the nation has benefited from the multi-billion-dollar event\end{tabular}         \\ \bottomrule                                                                                                                                                        
\end{tabular}
\caption{Additional generated sample from our model as in Table~\ref{tab:generated_example}.}
\vspace{-.1cm}
\label{tab:generated_example3}
\end{table*}

%
%
%
%Example 4
\begin{table*}[h]
\small
\centering
\begin{tabular}{ll}
\toprule
& Conversation 5                                                          \\ \toprule
Speaker                                                   & \begin{tabular}[c]{@{}l@{}}I’d take your point more seriously if it wasn’t based on animosity toward the umps. Let’s be more objective. \\

You’re daft if you think an automatic strike zone solves the issue. Technology isn’t even close to being ready to \\do that, no matter what Manfred says. It’s introduces many other problems as well.

Instant reply has already \\drastically slowed down the game\end{tabular}                                                                                                      \\ \midrule
Listener                                                   & \begin{tabular}[c]{@{}l@{}}Sure it is. Open up GameDay on your phone. It's 99.7\% accurate. All you need is 3 cameras and the computed \\can be almost perfect. It has been doing it since 1997. Look up pitchF\/X. Tech has been doing it forever. \\

The computer 99.7\% accuracy accounts for malfunctions. It's way better than 86-90\%\end{tabular} \\ \midrule

Speaker                                                    & \begin{tabular}[c]{@{}l@{}}’m not speaking only of accuracy. Having a computer call live pitches consistently is still far off. \\All it takes is a malfunction and your games get played under protest. 

Plus players attitudes won’t change. It won’t \\ fix anything.\end{tabular}                                                                                                                                                                                                                                             \\ \midrule
Listener                                                   & \begin{tabular}[c]{@{}l@{}}They have been calling live games consistently for 20 years. Have you ever watched a game on TV? \\The computer makes that strike zone. And it's 99.7\% accurate. 

Players attitudes wont change, but the game \\will be called fairly. So that's a plus.\end{tabular}                        \\ \midrule

Speaker                                                   & \begin{tabular}[c]{@{}l@{}}THAT strike zone is wrong often. Have you ever seen a game on TV? Lol.\end{tabular}                        \\ \midrule
Listener                                                   & \begin{tabular}[c]{@{}l@{}}Nope, it's not. It's the ump who is wrong.\end{tabular}                        \\ \midrule

Speaker                                                   & \begin{tabular}[c]{@{}l@{}}Yeah you’re right. Technology never malfunctions and works al of the time. How silly of me. \\My internet or electricity has never been out for days rendering everything useless.\end{tabular}                        \\ \midrule
Listener                                                   & \begin{tabular}[c]{@{}l@{}}Well, if you can find me umps who are 100\% perfect then I'd gladly support using them instead. I'll take \\ 99.7\% over 88\% anyday?\end{tabular}                        \\ \midrule

Speaker                                                   & \begin{tabular}[c]{@{}l@{}}88\% is never a number I’ve bet read about. And the point is not to be 100 perfect. No officiating crew is in any sport \\so why should baseball be different. 

The NFL allowing passing interference to be reviewable now is the\\  dumbest thing in the world. Technology is for enhancing strategic edge and assessment in sports, not for policing it.\end{tabular}                        \\ \midrule

Listener                                                   & \begin{tabular}[c]{@{}l@{}}So, one minute you say that robot umps should not come bacuse they arent 100\% perfect, and now you say that's not\\ the goal. Make up your mind, man! \url{https://tinyurl.com/jeuenvx}
This study lists their accuracy at 87\%. \\There are 100 others as well, Boston U and Yale have run tests of millions of pitches.

Technology is for ensuring\\ referee accuracy. And after last year's NFL season, they REALLY need it. I mean, FFS, the refs fixed the NFC \\ champsionship game so that NE could have an opponent for the super bowl who had no offense.\end{tabular}                        \\ \midrule

\begin{tabular}[c]{@{}l@{}}Speaker \\ (Model)\end{tabular} & \begin{tabular}[c]{@{}l@{}}This article says the right call is more important than tradition and that is what I’m arguing for. And 100\% \\ is not even remotely realistic. You want it to be right more than it is wrong.
Plus, how does a league actually control it\\ \\ 

\textcolor{blue}{Retrieved Knowledge 1:} There is a decrease in "bad" calls in this brief time span, suggesting umpires are striving \\ to make better calls. \\ \textcolor{blue}{Retrieved Knowledge 2:} Instant replay is being added to assist in making correct calls, since Major League Baseball \\made the decision that the right call is more important than tradition. \end{tabular}         \\ \bottomrule                                                                                                                                                        
\end{tabular}
\caption{Additional generated sample from our model as in Table~\ref{tab:generated_example}.}
\vspace{-.1cm}
\label{tab:generated_example4}
\end{table*}

\end{document}